\documentclass{article}

\usepackage[preprint,nonatbib]{neurips_2024}

\usepackage[utf8]{inputenc} 
\usepackage[T1]{fontenc}    
\usepackage{hyperref}       
\usepackage{url}            
\usepackage{booktabs}       
\usepackage{amsfonts}       
\usepackage{nicefrac}       
\usepackage{microtype}      
\usepackage{xcolor}         
\usepackage{amsmath} 
\usepackage{mathtools}
\usepackage{multirow}
\usepackage{bm}
\usepackage{arydshln}
\usepackage{subcaption}
\usepackage{wrapfig}

\title{FBI-LLM: Scaling Up Fully Binarized LLMs from Scratch via Autoregressive Distillation}

\author{
  Liqun Ma$^{1}$, ~Mingjie Sun$^{2}$,~Zhiqiang Shen$^{1}$ \\
  $^{1}$Mohamed bin Zayed University of AI\\ $^{2}$Carnegie Mellon University\\
  \texttt{\{Liqun.Ma,Zhiqiang.Shen\}@mbzuai.ac.ae}, \texttt{mingjies@andrew.cmu.edu} \\
}

\begin{document}

\maketitle

\begin{abstract}
  This work presents a {\bf F}ully {\bf BI}narized {\bf L}arge {\bf L}anguage {\bf M}odel (FBI-LLM), demonstrating for the first time how to train a large-scale binary language model from scratch (not the partial binary or ternary LLM like BitNet b1.58~\cite{ma2024era}) to match the performance of its full-precision counterparts (e.g., FP16 or BF16) in transformer-based LLMs. It achieves this by employing an autoregressive distillation (AD) loss with maintaining equivalent model dimensions (130M, 1.3B, 7B) and training data volume as regular LLM pretraining, while delivering competitive results in terms of perplexity and task-specific effectiveness. Intriguingly, by analyzing the training trajectory, we find that the pretrained weight is not necessary for training binarized LLMs from scratch. This research encourages a new computational framework and may facilitate the future design of specialized hardware tailored for fully 1-bit LLMs. We make all our models, code, and training dataset fully accessible and transparent to support further research\footnote{Code: \url{https://github.com/LiqunMa/FBI-LLM}. Model: \url{https://huggingface.co/LiqunMa/}.}. 
\end{abstract}

\section{Introduction}

\begin{wrapfigure}{r}{0.52\textwidth}
    \vspace{-0.18in}
    \centering
    \includegraphics[width=0.51\textwidth]{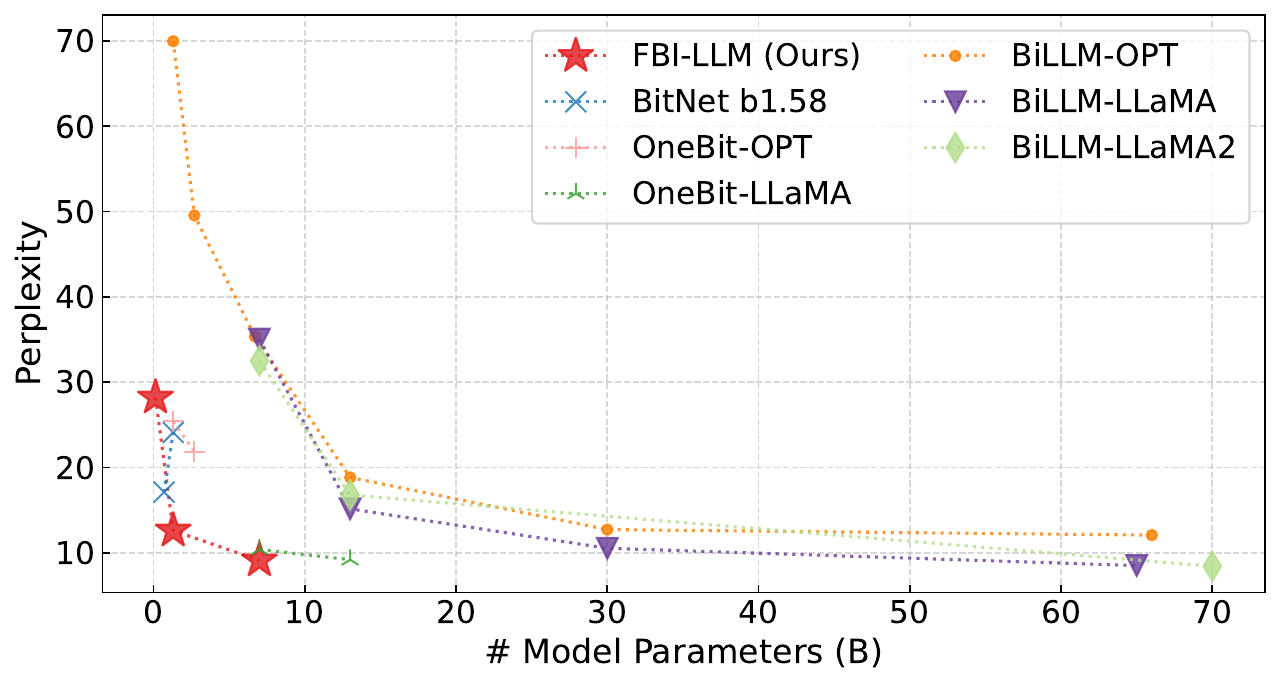}
    \vspace{-0.1in}
    \caption{Perplexity on Wikitext2 of existing binarized LLMs and our FBI-LLMs. FBI-LLMs get similar or lower magnitude of perplexity on similar size of models compared with other binarized LLMs.}
    \label{fig:compare}
    \vspace{-0.15in}
\end{wrapfigure}

Benefiting from the huge parameter scale and massive training corpora, transformer-based Large Language Models (LLMs), like ChatGPT~\cite{achiam2023gpt} and LLaMA~\cite{touvron2023llama,touvron2023llama2}, perform great in tasks requiring domain knowledge and complex reasoning. 
Moreover, the capabilities of LLMs tend to enhance as their parameter sizes expand. 
This substantial scale in parameters results in considerable storage and computational demands, which substantially restricts LLMs' broader application and development.
Quantization efficiently mitigates these limitations by mapping 32-bit parameters to smaller bit sizes. 
It substantially cuts storage requirements and enhances computational speed and energy efficiency during inference.

As the most extreme case of quantization, binarization represents each parameter by just \{-1, 1\}. 
It maximizes compression and efficiency but at the cost of accuracy. 
Prior efforts to preserve the efficacy of binarized LLMs include retaining salient parameters~\cite{shang2023pb} or using near-one-bit to represent each parameter~\cite{ma2024era}.
While these approaches have shown promise, they still leave room for optimization in storage and efficiency, and additional full-precision parameters or parameter encodings expressed in non-powers of 2 can cause extra overhead when adapting to edge hardware.

\begin{figure}[t]
    \centering
    \includegraphics[width=0.99\textwidth]{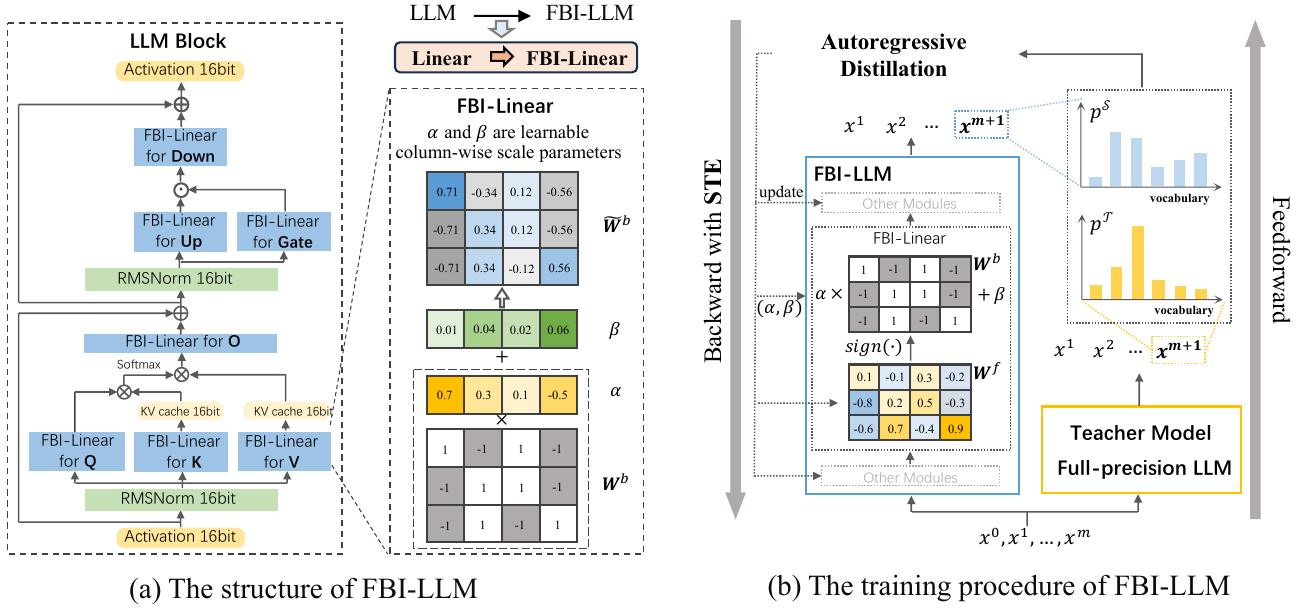}
    \caption{Illustration of the FBI-LLM framework. We take the structure of LLaMA as an example. {\bf Left}: the LLM block with the proposed FBI-Linear using learnable $\boldsymbol{\alpha}$ and $\boldsymbol{\beta}$. {\bf Right}: our autoregressive distillation and model pertaining procedure.}
    \label{fig:fbi-llm}
    \vspace{-0.2in}
\end{figure}

Some works on fully binarized LLMs are based on the optimization goal of minimizing the layer-wise $\ell_2$ loss~\cite{Frantar2022OptimalBC, shang2023pb} or performing binarization while continuing training a full-precision LLM with a small amount of data~\cite{xu2024onebit}. 
These methods face several issues: 
1) The binarization process greatly compresses the parameter space of the original model, damaging some of the knowledge stored in the full-precision model. 
Adequate training data is needed to allow the binarized model to relearn this knowledge and adapt it to the pattern of binarized parameters; 
2) Deriving binarized models from existing pretrained models does not allow for the selection of different parameter scales or vocabulary sizes, limiting their flexibility and practical application.

In this work, we propose a streamlined process for training \textbf{F}ully \textbf{BI}narized \textbf{LLMs} from scratch, termed \textbf{FBI-LLM}. 
To enable stable training of binary LLMs from scratch, we propose a novel training procedure based on distillation. 
Specifically, during training, we gradually distill from a full-precision teacher and adopt an autoregressive distillation-based scheme to match the predicted probabilities of the teacher model at each token location. 
With this simple autoregressive distillation loss, we can successfully train binarized LLMs from random initializations. 
Since our modifications are focused on the loss function, FBI-LLM can be easily incorporated into the existing LLM pre-training pipeline.  
Moreover, the binarization operation is decoupled from model training in this method, thus any techniques that enhance LLM training efficiency can be directly adapted for FBI-LLM. 

We empirically evaluate the effectiveness of our framework \textbf{FBI-LLM}, where we trained models with sizes ranging from 130M, 1.3B, to 7B. 
We use the widely-used Transformer architecture for LLMs, as can be seen in Fig.~\ref{fig:fbi-llm}. We show that we can train fully binarized LLMs from scratch, with a small performance gap as compared to full-precision counterparts.
Compared to baseline methods, our training process leads to fully binarized LLMs with better performance on perplexity (as shown in Fig. \ref{fig:compare}) and multiple downstream tasks.
We show that autoregressive distillation is key to training binarized LLMs. 
Further, analysis of pretraining checkpoints (e.g., flip-flop ratio and gradient norms) suggests there is no major difference between inheriting the weights from full-precision LLMs and training binarized LLMs from scratch.

Overall, the contribution of this paper can be summarized as follows: first, we demonstrate for the first time that we can successfully train LLMs with binary weights from scratch; second, we propose a novel loss formulation for stabilize the training of binarized LLMs, where we adopt autoregressive distillation to match the probability distribution of a teacher model; third, we conduct extensive experimental and analysis to demonstrate and better understand the effectiveness of our method.

\section{Related Work}

\noindent{\bf Neural Network Binarization}.  
Binarization, the most extreme form of network quantization, converts model parameters into a 1-bit format. 
Many studies have focused on Binary Neural Networks (BNNs) to improve their accuracy despite inherent limitations. 
BinaryConnect~\cite{binaryconnect} converts full-precision weights in neural networks to 1-bit binary weights using stochastic methods during training and simulates the effects of binary weights during inference.
They also implement a clipping function in backward propagation to prevent excessive growth of real-valued weights. 
Expanding on this, they develop the Binarized Neural Network (BNN)~\cite{hubara2016binarized}, which includes detailed training and acceleration techniques, demonstrating the efficiency and practicality of BNNs through reduced storage and faster processing times in image classification.

However, these methods typically suffer from accuracy loss, prompting numerous optimization-based solutions over recent years to mitigate this. 
Binary Weight Networks (BWN) and XNOR-Net~\cite{rastegari2016xnor} introduce a scaling factor that approximates floating-point parameters to reduce quantization errors. 
Further developments like DoReFa-Net~\cite{zhou2015dorefa}, WRPN~\cite{mishra2017wrpn}, and ABC-Net~\cite{lin2017abcnet} introduce strategies to minimize information loss and quantization errors. 
Innovations such as XNOR-Net++~\cite{rastegari2016xnor} and various mimic solutions like Distillation and Quantization (DQ)~\cite{polino2018disquant} continue to refine these approaches, emphasizing stability and high accuracy in training binary models. 
To address the non-differentiability of the binarization function, techniques like the straight-through estimator (STE)~\cite{bengio2013estimating} are used for backpropagation. 
The ReActNet~\cite{liu2020reactnet} improves BNNs with generalized activation functions, and the BNN+~\cite{darabi2019bnn} introduces an enhanced derivative approximation of the sign function along with a regularization strategy to optimize weight learning.

\noindent{\bf Large Language Model Binarization}.  
PB-LLM~\cite{shang2023pb} implements partial binarization of the LLMs, retaining the salient parameters at full precision, occupying only a small portion of all parameters, to maintain the linguistic reasoning capacity. 
BiLLM~\cite{huang2024billm} also considers the distribution pattern of the weight scale. 
It uses a binary residual approximation strategy to binarize salient parameters, which consists of an original binary tensor and a residual binarized matrix to present the binarization result of salient parameters.
BitNet b1.58~\cite{ma2024era} quantizes all parameters to the set of \{-1, 0, 1\}, where they find that quantized LLMs achieve competitive performance to their full-precision counterparts.
However, PB-LLM and BiLLM use the extra cost of storage to process salient weights, and BitNet b1.58 uses an average of 1.58 bits to represent weights.
None of them have reached the limits of binary models. 
There is room for further improvement in model storage size and inference speed. 
Our work focuses on achieving fully binarized LLMs while preserving the model's capabilities as much as possible.

BitNet \cite{wang2023bitnet} and OneBit~\cite{xu2024onebit} employ quantization-aware training (QAT) to binarize LLMs. 
BitNet utilizes group quantization by applying different scales to the parameters of various groups, which accelerates model training. Its training loss aligns with that of pretraining autoregressive language models. Conversely, OneBit preserves the full-precision model knowledge through quantization-aware knowledge distillation, using a full-precision model as the teacher and guiding the binarized model training with two distinct loss functions. Unlike these two methods, our approach achieves similar or better results through a more streamlined and efficient training process.

\vspace{-0.05in}
\section{Methodology}
\label{method}
\vspace{-0.05in}
In this section, we first provide an overview of the architecture of our FBI-LLM in Section \ref{arch}. Then, in Section \ref{linear}, we detail the FBI-linear module, the main component of FBI-LLM. Finally, we elaborate the FBI-LLM autoregressive distillation-based training procedure in Section \ref{train}.

\subsection{Architecture of FBI-LLM}
\label{arch}
We illustrate the overall architecture of FBI-LLM in Fig.~\ref{fig:fbi-llm} (a).
In transformer-based LLMs, the majority of parameters are found within the linear modules.
FBI-LM replaces all linear modules, except for the causal head, with FBI-linear (Fully BInarized Linear).
Since the causal head directly influences the output token distribution in each step, binarizing its parameters would significantly affect the accuracy of the model's output, so we retain its precision. 

Additionally, the parameters in two other core modules of LLMs, embedding and layer norm, also need to be kept at full precision. 
This is because the embedding module contains semantic information about all tokens and, as the first layer of the model input, determines the text's initial representation. 
Layernorm, on the other hand, scales the activation values directly. 
Binarizing its parameters would significantly reduce the semantic expressiveness of the activation values at each layer. 
Similar settings are also adopted in other work~\cite{wang2023bitnet} about the binarization of LLMs.

\subsection{FBI-Linear}
\label{linear}
The main parameters in FBI-linear are a matrix $\boldsymbol{W}^b \in\mathbb{R}^{m\times n}$ consisting only of $\{1, -1\}$, which is binarized from the full-precision linear module $\boldsymbol{W}^f \in\mathbb{R}^{m\times n}$ of the LLMs. 
During the training, the binarization process can be formulated as:
\begin{equation}
    \label{bi_1}
    \boldsymbol{W}^b = sign(\boldsymbol{W}^f)
\end{equation}
where $sign$ function is formulated as: 
\begin{equation}
\label{bi_2}
sign(\boldsymbol{W}_{ij}^f)=
\begin{dcases}
  1, & \boldsymbol{W}_{ij}^f > 0 \\
  -1, & \boldsymbol{W}_{ij}^f \leq 0
\end{dcases}
\end{equation}

We follow the previous works~\cite{rastegari2016xnor, wang2023bitnet} to scale the binarized parameter with full-precision scale factors.
Scale factors can effectively reduce the error between the binarized and original parameters, thereby preserving the more representational capacity of the corresponding module. 
They constitute a small fraction of the total parameters, making them a highly efficient enhancement to the model's performance without significantly increasing the parameter count and computational demand.

Specifically, in the FBI-linear, we apply scaling at the granularity of the matrix columns. 
The calculation process can be formulated as:
\begin{equation}
    \label{fbi-linear}
    \widetilde{\boldsymbol{W}}^b_{\cdot,j} = \alpha_j\boldsymbol{W}^b_{\cdot,j}+\beta_j
\end{equation}
where $\widetilde{\boldsymbol{W}}^b_{\cdot,j}$ donate the $j^{th}$ column of the scaled binarized weight matrix $\widetilde{\boldsymbol{W}}^b\in\mathbb{R}^{m\times n}$. 
$\alpha_j$ and $\beta_j$ are the $j^{th}$ element in learnable scale vectors $\boldsymbol{\alpha} \in \mathbb{R}^n$ and $\boldsymbol{\beta} \in \mathbb{R}^n$ respectively.

To accelerate the model's convergence speed, we initialize $\boldsymbol{\alpha}$ and $\boldsymbol{\beta}$ before training as following:
\begin{equation}
    \label{init_alpha}
    \alpha_j = a_j 
\end{equation}

\begin{equation}
    \label{init_beta}
    \beta_j = \frac{1}{m}\sum^m_i|\boldsymbol{W}_{ij}^f-a_j|
\end{equation}
where $a_j=\frac{1}{m}\sum^m_i\boldsymbol{W}_{ij}^f$, denotes the average of $j^{th}$ column in $\boldsymbol{W}^f$.

\subsection{Autoregressive Distillation} \label{train}

Given a training corpus of tokens ${\bm x}=\left\{x_1, \ldots, x_n\right\}$, a standard autoregressive language modeling objective~\cite{radford2018improving} is to maximize the likelihood:
\begin{equation}
\mathcal L({\bm x})=\sum_i \log p\left(x_i \mid x_{i-k}, \ldots, x_{i-1} ; \boldsymbol{\theta}\right)
\end{equation}
where  $k$ represents the size of the context window and the conditional probability $p$ is modeled through a neural network characterized by the parameters $\boldsymbol{\theta}$. Unlike conventional autoregressive language models, we train FBI-LLM using \textbf{autoregressive distillation (AD)}. 
In the training, a full-precision pre-trained LLM is used as the teacher model, and the binarized target model acts as the student. Suppose each instance of training data consists of a sequence of input tokens $x^1, \dots, x^m$, the teacher prediction probability for the next token can be formulated as:
\begin{equation}
\boldsymbol{p}^{\mathcal{T}}\left(x^{m+1} \mid x^1, \ldots, x^m\right)=\operatorname{softmax}\left(\boldsymbol{h}_l^m \boldsymbol{W}_{m+1}\right) .
\end{equation}
where $\boldsymbol{h}_l^m$ represents the activation of the final transformer block, $\boldsymbol{W}_{m+1}$ represents parameters of the added linear output layer to predict the next token's probability.

The cross-entropy between the outputs of the student model and the teacher model is calculated as the final loss function at each step of predicting the next token. 
It can be formulated as:
\begin{equation}
    \label{loss}
    \mathcal{L} = -\frac{1}{n}\sum^n_{i} \boldsymbol{p}^{\mathcal{T}}(x^{i+1}) \cdot \log \boldsymbol{p}^{\mathcal{S}}(x^{i+1})
\end{equation}
Here $n$ denotes the number of input tokens. 
$\boldsymbol{p}_j^{\mathcal{T}}(x^{i+1})$ denotes the token distribution over the vocabulary at the $i^{th}$ step predicted by the teacher model, while $\boldsymbol{p}^{\mathcal{S}}(x^{i+1})$ is the corresponding predicted distribution of the student model.

QAT utilizing knowledge distillation has been shown to be effective in various studies~\cite{kim2019qkd,boo2021stochastic,pham2023collaborative,bhardwaj2024oh,xu2024onebit}. However, unlike these works, our training process exclusively uses the autoregressive distillation loss without adding any other losses to maintain simplicity. Our experiments verified that using only the distillation loss yields better results than the vanilla one-hot label based autoregressive loss while maintaining methodological simplicity when working with fully binarized LLMs.

Since the $sign(\cdot)$ is non-differentiable at zero, it causes the gradient chain to break during backpropagation, preventing optimization of the model parameters. 
Therefore, we use the Straight-Through Estimator (STE) method~\cite{bengio2013estimating} during backpropagation, where the gradient of the output of the non-differentiable function is used as an estimate for the gradient of the input, thus allowing the gradient to be effectively propagated.
This estimation can be formulated as:
\begin{equation}
    \label{ste}
    \frac{\partial \mathcal{L}}{\partial \boldsymbol{W}^f} = \frac{\partial \mathcal{L}}{\partial \boldsymbol{W}^b}
\end{equation}

\vspace{-0.1in}
\section{Experiments}\label{experiments}

In our experiment, we follow the W1A16 setup~\cite{huang2024billm,xu2024onebit}, quantizing only the parameters to 1-bit while keeping the activation values at 16-bit. We train FBI-LLMs with sizes of 130M, 1.3B, and 7B, testing their performance across multiple tasks.

\subsection{Setup}
\textbf{Dataset.}  
We train FBI-LLMs with the Amber dataset~\cite{liu2023llm360}. 
Amber dataset is a mixture of RefinedWeb~\cite{penedo2023refinedweb}, StarCoder~\cite{li2023starcoder}, and RedPajama-v1~\cite{together2023redpajama} and contains the total 1.26 trillion tokens.
It divides the data into 360 chunks, with each chunk containing an average of 3.5 billion tokens\footnote{As shown in Fig.~\ref{fig:7b}, in our experiment, about 10\% of the training data chunks have already achieved competitive performance. Further training is naturally expected to yield even higher accuracy.}.

\textbf{Training details.}  
Our models used for experiments adopt a similar structure as LLaMA2~\cite{touvron2023llama}. 
For the specific hyper-parameters settings of FBI-LLMs of different sizes, refer to Table \ref{tab:hypara}.
The maximum sequence length is set to 2048. 
The optimizer is Adam with $\beta_1 = 0.9$, $\beta_2 = 0.98$. 
The initial learning rate for all model sizes is set at $3e-4$, following a cosine learning rate schedule that decreases to a final rate of $3e-5$ as it is warmed up over 2,000 steps.
We use gradient clipping at 1.0.
We use LLaMA2-7B as the teacher model for all size FBI-LLMs to calculate autoregressive distillation loss.
We train models with 64 NVIDIA A100 GPUs in total and maintain BF16 precision while training. Please refer to Appendix~\ref{sec:confanddetail} for more details.

\begin{table}
  \caption{Hyper-parameters for FBI-LLMs in xxperiments.}
  \label{tab:hypara}
  \centering
  \scalebox{0.9}{
  \begin{tabular}{lccccc}
    \toprule
    Model Size    & \# layers & hidden size & \# attention heads		&intermediate size 	&batch size (tokens)\\
    \midrule
    FBI-LLM 130M  & 12 & 768  & 12  &  2,048  & 2M  \\
    
    FBI-LLM 1.3B  & 24 & 2,048 & 32  &  5,632  & 2.4M   \\
    
    FBI-LLM 7B    & 32 & 4,096 & 32  &  11,008 & 3.9M\\

    \bottomrule
  \end{tabular}
  }
  \vspace{-0.15in}
\end{table}

\textbf{Baselines.} 
We compare our work with prior binarized LLMs Bi-LLM~\cite{huang2024billm}, OneBit~\cite{xu2024onebit}, and BitNet~\cite{wang2023bitnet}. 
We also include the BitNet b1.58~\cite{ma2024era}, which is a ternary quantization LLM, as our baseline model for comparison\footnote{As the original BitNet b1.58~\cite{ma2024era} has not open-sourced their model, we use a third-party open-sourced one \url{https://huggingface.co/1bitLLM/} to evaluate certain indicators for the comparison. This model achieves slightly better results than those reported in the original paper.}. 
We further include results from open-sourced full-precision models of various sizes, such as OPT~\cite{zhang2022opt}, LLaMA~\cite{touvron2023llama, touvron2023llama2}, and TinyLLaMA~\cite{zhang2024tinyllama}, as references.

\textbf{Evaluation Metrics.} 
We evaluate the models based on their zero-shot performance in some downstream tasks, including BoolQ~\cite{Clark2019BoolQET}, PIQA~\cite{bisk2020piqa}, HellaSwag~\cite{Zellers2019HellaSwagCA}, Winogrande \cite{sakaguchi2021winogrande}, ARC~\cite{Clark2018ThinkYH}, and OpenbookQA~\cite{Mihaylov2018CanAS}.
We also use perplexity as the evaluation metric. 
Perplexity measures how well a probability model predicts a token, quantitatively measuring the model’s generation power.

\subsection{Main Results}

Table \ref{tab:main} presents the main results comparing our FBI-LLMs to various state-of-the-art baseline models. We also report the average bit-width occupied by model parameters, excluding the embedding layer and the head, for different models. Details on the calculation process can be found in Appendix \ref{sec:bw}. Our FBI-LLMs maintain the lowest average bit-width across different model sizes while demonstrating remarkable performance. We provide zero-shot accuracy, which is a foundation for understanding how well a model can perform without additional task-specific information. This metric is commonly used to assess the model's initial capabilities and aligns with certain benchmarking tasks aimed at measuring the pre-trained LLM's general comprehension and knowledge-reserving capabilities across diverse downstream tasks without additional few-shot information.

Since there is no binary baseline for the 130M size, we compare our 130M model with the BitNet b1.58 at the 700M scale. Despite the fivefold difference in model size and significant variations in quantization degree, our model still outperforms BitNet b1.58 in BoolQA and OpenbookQA. 
For the 1.3B-scale binary models, our FBI-LLM achieves the best performance in most downstream tasks and perplexity, even matching or exceeding the capacity of some 7B-scale binary models like BiLLM-LLaMA2-7B. Compared to the full-precision models of a similar scale, the proposed FBI-LLM 1.3B can achieve up to 87\% of their performance in downstream tasks. 
In the 7B scale, our model significantly outperformed nearly all baselines.

\begin{wrapfigure}{r}{0.4\textwidth} 
  \centering
  \vspace{-0.2in}
  \includegraphics[width=0.4\textwidth]{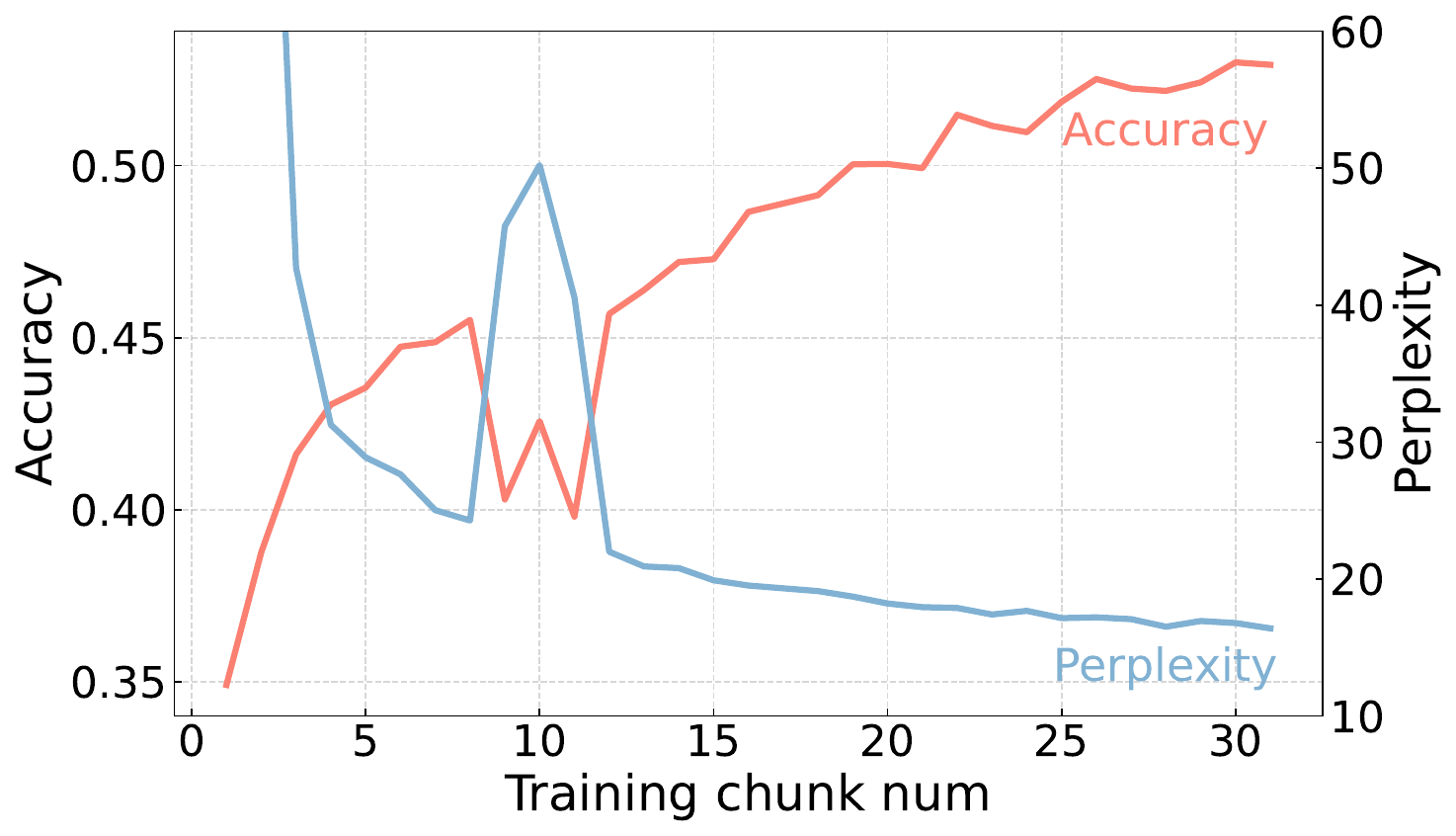} 
  \vspace{-0.25in}
  \caption{Changes in average perplexity and downstream task accuracy during the training of FBI-LLM 7B. The horizontal axis represents the number of Amber data blocks used for training.}
  \label{fig:7b}
  \vspace{-0.35in}
\end{wrapfigure}
Furthermore, limited by computational resources, the current results for FBI-LLM 7B are not final. We only use 8.6\% (31 chunks) of the Amber dataset. Fig. \ref{fig:7b} illustrates the changes in downstream task accuracy and perplexity during the training process of FBI-LLM-7B. It is clear that, as of the current training progress, the performance of FBI-LLM-7B will be improved consistently, and further training could yield better results.

\begin{table}
  \caption{Performance on downstream tasks and perplexity. Here, BW means bit-width, which refers to the average number of bits occupied by each parameter. HS, WG, and OBQA are abbreviations for HellaSwag, Winogrande, and OpenbookQA, respectively. We divide the table into three blocks based on model size. In each block, the bold values represent the best values among the non-high-precision models, while the values with an underline represent the best values among the high-precision models.}
  \label{tab:main}
  \centering
  \scalebox{0.71}{
  \begin{tabular}{l|cc|cccccccc|ccc}
    \toprule
    \multirow{2}{*}{Model}  &\multirow{2}{*}{Size} &\multirow{2}{*}{BW}  &\multicolumn{8}{c}{Zero-shot Accuracy  $\uparrow$}  &\multicolumn{3}{|c}{Perplexity $\downarrow$} \\ 
    &           &           &BoolQ  &PIQA  &HS    &WG    &ARC-e  &ARC-c  &OBQA  &Ave.  &Wiki2  &PTB     &C4   \\
    \midrule
    BitNet b1.58~\cite{ma2024era}       &700M &1.59       &58.2   &\textbf{68.1}  &\textbf{35.1}  &\textbf{55.2}  &\textbf{51.8}   &\textbf{21.4}   &20.0  &\textbf{44.3}  &\textbf{17.1}   &\textbf{72.1}    &\textbf{17.5}    \\
    FBI-LLM (Ours)                         &130M &1.01   &\textbf{62.1}   &59.3  &28.7  &51.0  &34.9   &20.5   &\textbf{26.4}  &40.4      &28.2   &136.6   &26.9 \\
    \hline
    TinyLLaMA~\cite{zhang2024tinyllama} &1.1B &16 &\underline{57.8} &\underline{73.3} &\underline{59.2} &59.1            &\underline{55.3} &\underline{30.1} &\underline{36.0} &\underline{53.0} &\underline{7.8}    &30.5    &\underline{9.9}  \\
    OPT~\cite{zhang2022opt}             &1.3B &16     &\underline{57.8}   &72.5              &53.7              &\underline{59.5}  &51.0               &29.5   &33.4  &51.1  &14.6   &\underline{20.3}    &16.1 \\
    \hdashline
    OneBit-OPT~\cite{xu2024onebit}      &1.3B &1.02       &59.5          &62.6           &34.3          &51.1          &41.3          &24.1            & -    &-     &25.4   &-       &23.0 \\
    BitNet b1.58~\cite{ma2024era}       &1.3B &1.59       &56.7          &68.8           &37.7          &\textbf{55.8} &\textbf{54.9} &24.2            &19.6  &45.4  &24.1   &145.1   &21.8    \\
    FBI-LLM (Ours)                        &1.3B &1.01   &\textbf{60.3} &\textbf{69.0}  &\textbf{42.3} &54.0          &43.6          &\textbf{25.3}   &\textbf{29.6}  &\textbf{46.3}  &\textbf{12.6}   &\textbf{39.3}    &\textbf{13.8} \\
    \hline
    OPT~\cite{zhang2022opt}             &7B   &16     &66.1               &76.5              &67.2  &65.4  &60.0   &34.7   &37.4  &58.2  &10.9   &\underline{15.8}    &12.7 \\
    LLaMA~\cite{touvron2023llama}       &7B   &16     &75.1               &\underline{79.2}  &\underline{76.2}  &\underline{69.9}  &72.9   &44.9   &\underline{44.4}  &66.0  &5.7    &41.2    &\underline{7.3} \\
    LLaMA2~\cite{touvron2023llama2}     &7B   &16     &\underline{77.7}   &79.1              &76.0  &69.1  &\underline{74.6}   &\underline{46.2}   &44.2  &\underline{66.7}  &\underline{5.5}    &37.9    &\underline{7.3}  \\
    \hdashline
    OneBit-LLaMA2~\cite{xu2024onebit}                     &7B   &       &\textbf{63.1}   &68.1  &52.6  &58.4  &41.6   &29.6   & -    &-          &9.7    &-       &11.1 \\  %
    BitNet~\cite{wang2023bitnet}        &7B   &-      &-   &-     &38.9  &51.4     &-      &-      &-     &-      & -     &-       & -   \\
    BiLLM-OPT~\cite{huang2024billm}     &7B   &1.11   &62.2   &58.6  &31.9  &51.5  &34.1   &23.9   &29.0  &41.6  &35.4   &73.6    &43.2 \\  
    
    BiLLM-LLaMA~\cite{huang2024billm}  &7B   &1.08   &62.7   &61.2  &36.8  &51.1  &36.0   &25.7   &31.8  &43.6  &35.0   &421.3  &39.6 \\  
    
    BiLLM-LLaMA2~\cite{huang2024billm}  &7B   &1.08   &61.8   &60.6  &34.8  &52.4  &36.2   &24.4   &{33.2}  &43.3  &32.5   &3877.4  &40.5 \\  
    FBI-LLM (Ours)                            &7B   &1.01   &61.5   &\textbf{72.6}  &\textbf{57.7}  &\textbf{58.9}  &\textbf{53.0}   &\textbf{29.9 }  &\textbf{36.8}  &\textbf{52.9}      &\textbf{9.1}   &\textbf{29.6}    &\textbf{10.5} \\
    \bottomrule
  \end{tabular}
}
\end{table}

\subsection{Effectiveness of Autoregressive Distillation}
\label{sec:ead}
\begin{figure}[htbp]
\centering
    \begin{subfigure}[b]{0.49\textwidth}
    \centering
    \includegraphics[width=\textwidth]{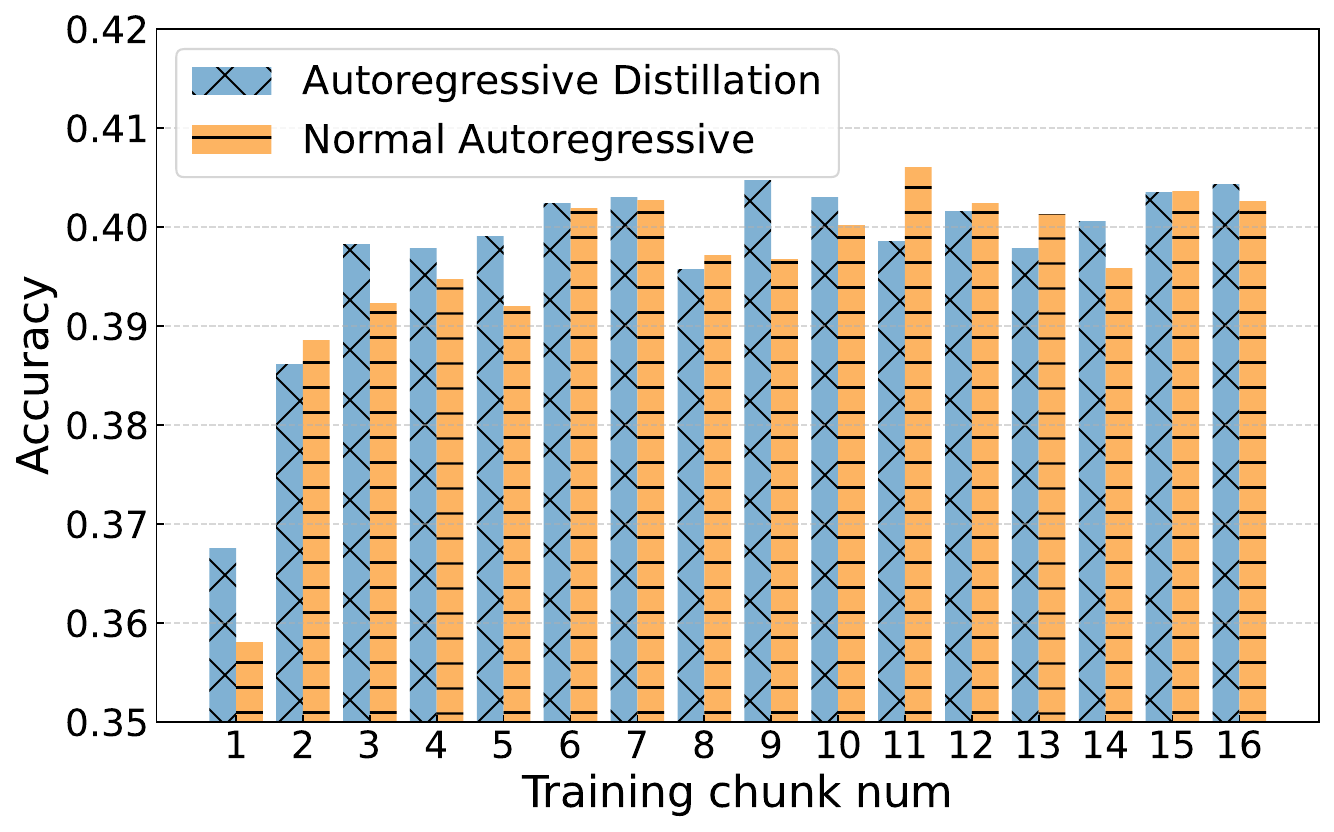}
    \caption{Average downstream task accuracy}
    \end{subfigure}
    \begin{subfigure}[b]{0.49\textwidth}
    \centering
    \includegraphics[width=\textwidth]{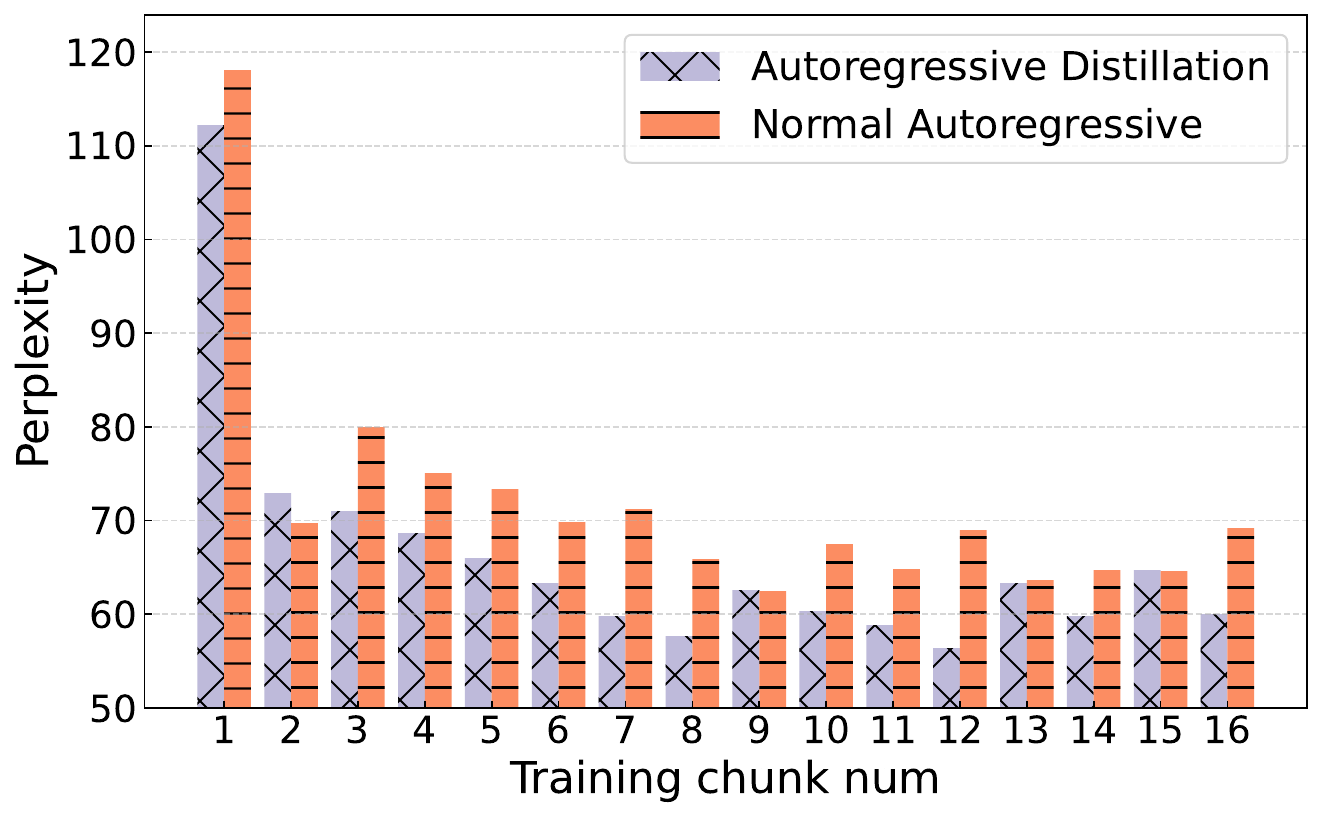}
    \caption{Average perplexity}
    \end{subfigure}
    \caption{The model performance for different training loss.}
    \label{fig:ab}
    \vspace{-0.2in}
\end{figure}

To demonstrate the effectiveness of using only autoregressive distillation as the training objective, we train two models: one using solely the autoregressive distillation loss and the other using only the standard autoregressive loss. All other training procedures are identical to those used for FBI-LLM. We evaluate the performance of these models on downstream tasks and perplexity, as shown in Fig.~\ref{fig:ab}. The evaluation tasks and datasets are the same as those listed in Table~\ref{tab:main}. For clarity, we present only the average accuracy across different tasks and the average perplexity across different datasets here. Detailed performance for each task is provided in Appendix~\ref{sec:detail_exp}.
 
It can be observed that throughout the training process, models trained with autoregressive distillation objective consistently outperform those trained with the standard autoregressive scheme in downstream tasks and perplexity. 
This indicates that using autoregressive distillation objective is more effective in training binarized LLMs. The utilized soft labels from the output probabilities of the teacher model contain more information than hard labels (i.e., the vocabulary labels). They provide a distribution over all possible vocabulary, indicating not just the target word but also the relative confidence in other possible words. This richer information helps the target model learn nuanced patterns and relationships in the data that are captured by the strong teacher model. Since our target model is learning solely from a smoothed version of the ground truth, it is less likely to overfit to the noise or specific details in the training data that may not generalize well to new data. 
Therefore, retaining only autoregressive distillation as the training objective ensures the simplicity and effectiveness of the entire training workflow.

\section{Analysis}

In this section, we analyze 1) the better choice of training from scratch or continuing training from pretrained LLM for binarized LLMs. 2) training instability and our solution. 3) storage efficiency of our models. 4) generation case demonstrations.

\subsection{Training from Scratch or Continue Training from Pretrained LLM?}

\begin{figure}[htbp]
\centering
    \begin{subfigure}[b]{0.493\textwidth}
        \centering
        \includegraphics[width=\textwidth]{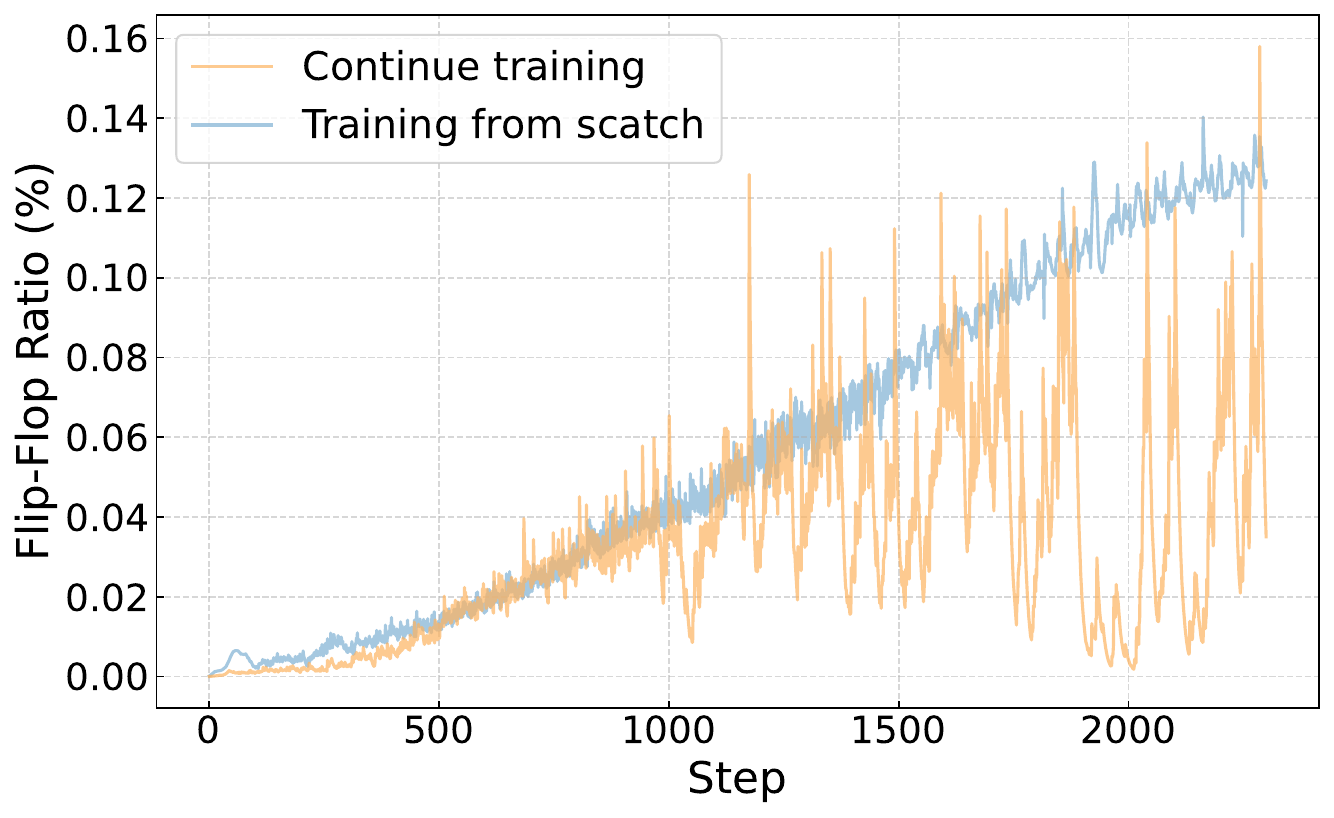}
        \caption{Average flip-flop ratio}
        \label{fig:ffr}
    \end{subfigure}
    \begin{subfigure}[b]{0.485\textwidth}
        \centering
        \includegraphics[width=\textwidth]{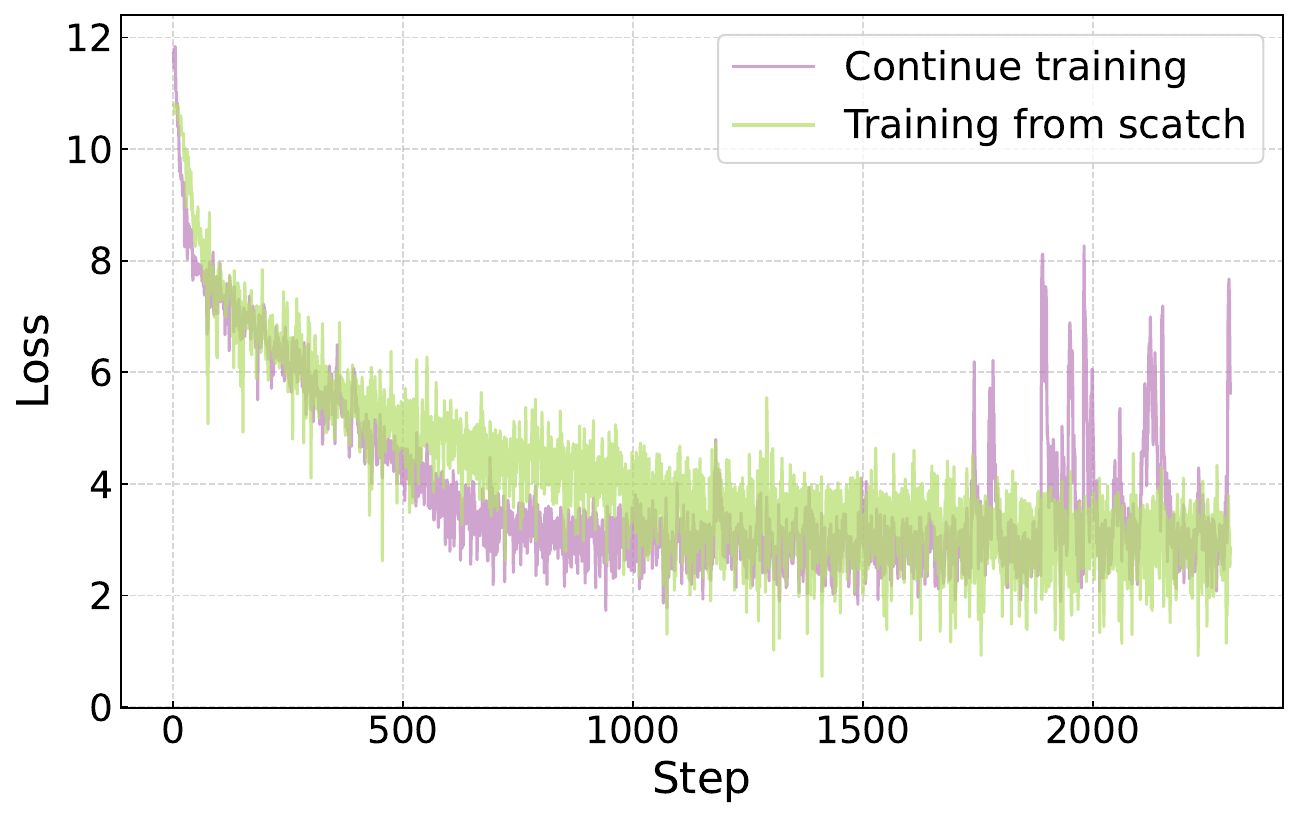}
        \caption{Training loss}
        \label{fig:loss}
    \end{subfigure}
    \caption{The flip-flop ratio and loss for different training procedures.}
\vspace{-0.1in}
\end{figure}

Intuitively, continuing training from a pretrained LLM to obtain a binarized model can inherit knowledge from the original model, potentially achieving better results than training from scratch. 
To evaluate this hypothesis, we conduct analytical experiments to record and compare the behaviors of models under both training procedures.

To quantify and examine the model behaviors using pretrained parameters or training from scratch, as well as their stability and initialization dependency of them, we apply the flip-flop ({\bf FF}) ratio~\cite{liu2021adam}. The FF ratio measures optimization stability behavior, which is defined as follows:
\begin{equation}
\begin{aligned}
& \mathbf{C}_{\mathbf{FF}}=\frac{\left|{Sign}\left(\boldsymbol{w}^b_{t+1}\right)-{Sign}\left(\boldsymbol{w}^b_t\right)\right|_{\text{abs}}}{2},
\end{aligned}
\end{equation}
where $\mathbf{C}_{\mathbf{FF}}$ is an indicator that shows whether a binary weight changes its sign after the update at iteration \(t\), and $|.|_\text{abs}$ is the absolute operation. 
\begin{equation}
\begin{aligned}
& \mathbf{FF}_{\text {ratio}}=\frac{\sum_{l=1}^L \sum_{\boldsymbol{w}^b \in \boldsymbol{W}_l^b} \mathbf{C}_{\mathbf{FF}}}{N_{\text {total}}},
\end{aligned}
\end{equation}
where $N_{\text{total}}$ represents the total number of parameters in a model with $L$ layers. \(\mathbf{FF}_{\text{ratio}}\) denotes the flip-flop ratio, which is the percentage of parameters that change their signs. 

In this experiment, we select TinyLLaMA as the base LLM and initialize the binarized LLM parameters using either pretrained values or random assignments.
We maintain consistency in all other training details with the FBI-LLM. 
In addition to the \(\mathbf{FF}_{\text{ratio}}\), we monitor training losses and gradient norms prior to gradient clipping across both training procedures.

\begin{wrapfigure}{r}{0.4\textwidth} 
  \centering
  \vspace{-0.2in}
  \includegraphics[width=0.4\textwidth]{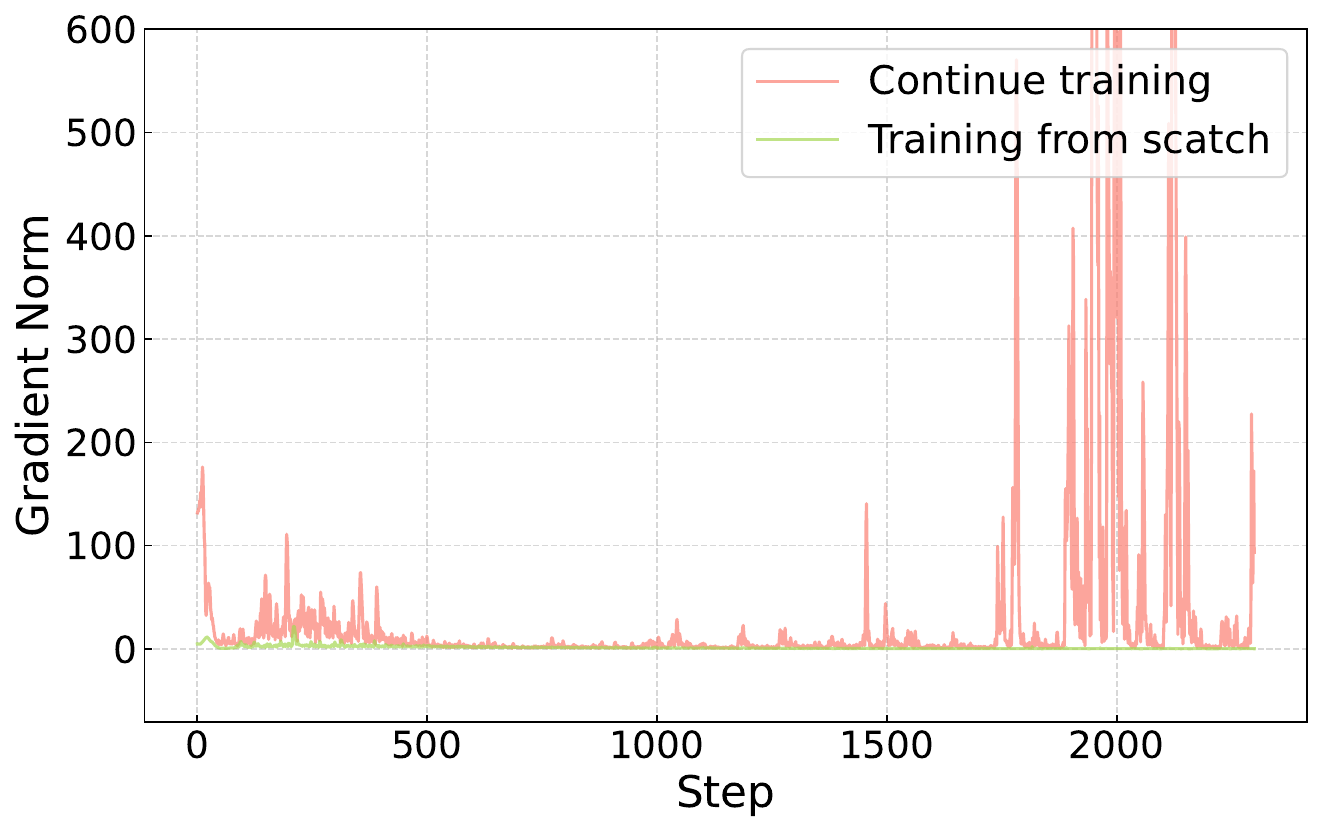} 
  \vspace{-0.25in}
  \caption{Gradient norm curves from different training procedures.}
  \label{fig:norm}
  \vspace{-0.25in}
\end{wrapfigure}

From Fig.~\ref{fig:ffr}, it is observed that during the beginning of training, the trend of \(\mathbf{FF}_{\text{ratio}}\) remains largely consistent between the two procedures.
Throughout the training process, the \(\mathbf{FF}_{\text{ratio}}\) for both methods are on similar scales and the values are relatively small.
It indicates that the two different parameter initialization approach do not significantly differ in their effects on binarization optimization process. 
Fig.~\ref{fig:loss} shows changes in training loss for two training procedure.
At the initial stage of training, the training losses for the two approaches are essentially the same, shows that the model's training loss is not notably influenced by the initialization approach.
Although the loss of training from scratch is slightly higher than that of continuing training after a period of time, after some more time, the loss of training from scratch once again matches or even becomes less than that of continuing training, and the trend of loss change in training from scratch is more stable. Significantly, at approximately the $1000^{th}$ step, as illustrated in Fig.~\ref{fig:ffr}, the \(\mathbf{FF}_{\text{ratio}}\) begins to show notable fluctuations when continuing training from a pretrained LLM.
Similarly, at around the $1700^{th}$ step shown in Fig.~\ref{fig:loss}, the training loss experiences similar issues.
Additionally, Fig.~\ref{fig:norm} highlights more pronounced changes in the gradient norms during the continuing training.

These findings challenge our initial hypothesis that starting with a pretrained LLM would endow the binarized model with inherited knowledge, thus enhancing performance. 
Instead, they imply that binarization through training is not sensitive to the way of parameter initialization. 
Furthermore, we speculate that binarized and full-precision LLMs employ different parameter combinations and configurations to encode semantics, which results in substantial divergences in their parameter space pattern.
To adapt this pattern, the optimization process for binarization by continuing training from a pretrained LLM might necessitate more profound parameter adjustments.
This could partly explain why it is more unstable compared to training from scratch during the training.

\subsection{Training Instability}
Both binary and full-precision LLM training have been found to exhibit unstable training behaviors~\cite{wang2023bitnet, xu2024onebit, chowdhery2023palm}. 
Our FBI-LLM exhibits similar issues, specifically manifesting as sudden spikes in training loss when training 1.3B and 7B FBI-LLMs, which sometimes fail to converge after that. 
We adopt the solution similar to PaLM~\cite{chowdhery2023palm}: if the loss no longer tends to converge, the model reverts to the previous checkpoint and skips the data chunk that triggered the unstable loss to continue training. 
The model no longer encounters issues at the same training steps using this approach. 
We observe that pretraining the 7B FBI model from scratch has approximately a 6\% probability of causing loss spikes. For the 1.3B model, training is more unstable due to the lower capability, with about a 15\% probability of loss spikes. This is consistent with the pretraining behavior seen in real-valued LLMs while the probability of spiking is significantly higher, which may be related to the limited expressive capability of binary parameters. To handle this, we skip any data blocks where loss spikes occur.

\subsection{Storage Efficiency}
\begin{table}[h]
  \vspace{-0.15in}
  \caption{Compression and extra parameters ratio.}
  \label{tab:storage}
  \centering
  \scalebox{0.82}{
  \begin{tabular}{lcccc}
    \toprule
    \textbf{Model}      &\textbf{Model Size}          &\textbf{Storage Size} &\textbf{Compression Ratio} &\textbf{Extra Parameter Ratio } \\
    \midrule
    LLaMA         &130M        &0.25GB  &\multirow{2}{*}{59.26\%}  &\multirow{2}{*}{0.119\%}  \\
    FBI-LLM       &130M        &0.10GB  &    &\\
    \midrule
    LLaMA         &1.3B        &2.54GB  &\multirow{2}{*}{84.67\%}  &\multirow{2}{*}{0.063\%} \\
    FBI-LLM       &1.3B        &0.39GB  &    &\\
    \midrule
    LLaMA         &7B          &12.55GB &\multirow{2}{*}{90.07\%}  &\multirow{2}{*}{0.034\%} \\
    FBI-LLM       &7B          &0.39GB  &    & \\
    \bottomrule
  \end{tabular}}
\end{table}

Table~\ref{tab:storage} presents the theoretical storage space required by FBI-LLMs of various sizes compared to the full-precision LLaMA with the same structure. It also details the proportion of additional parameters ($\boldsymbol{\alpha}$ and $\boldsymbol{\beta}$) introduced by FBI-LLM. The comparison in the table demonstrates that FBI-LLM can achieve a high compression ratio, significantly reducing the storage burden of LLMs. Although the extra parameters for scaling and shifting introduced by FBI-LLM need to be retained in full precision, their proportion is minimal, rendering their impact on storage negligible.

\subsection{Generation Cases}
\label{sec:case}
\begin{figure}[ht]
    \centering
    \includegraphics[width=0.96\textwidth]{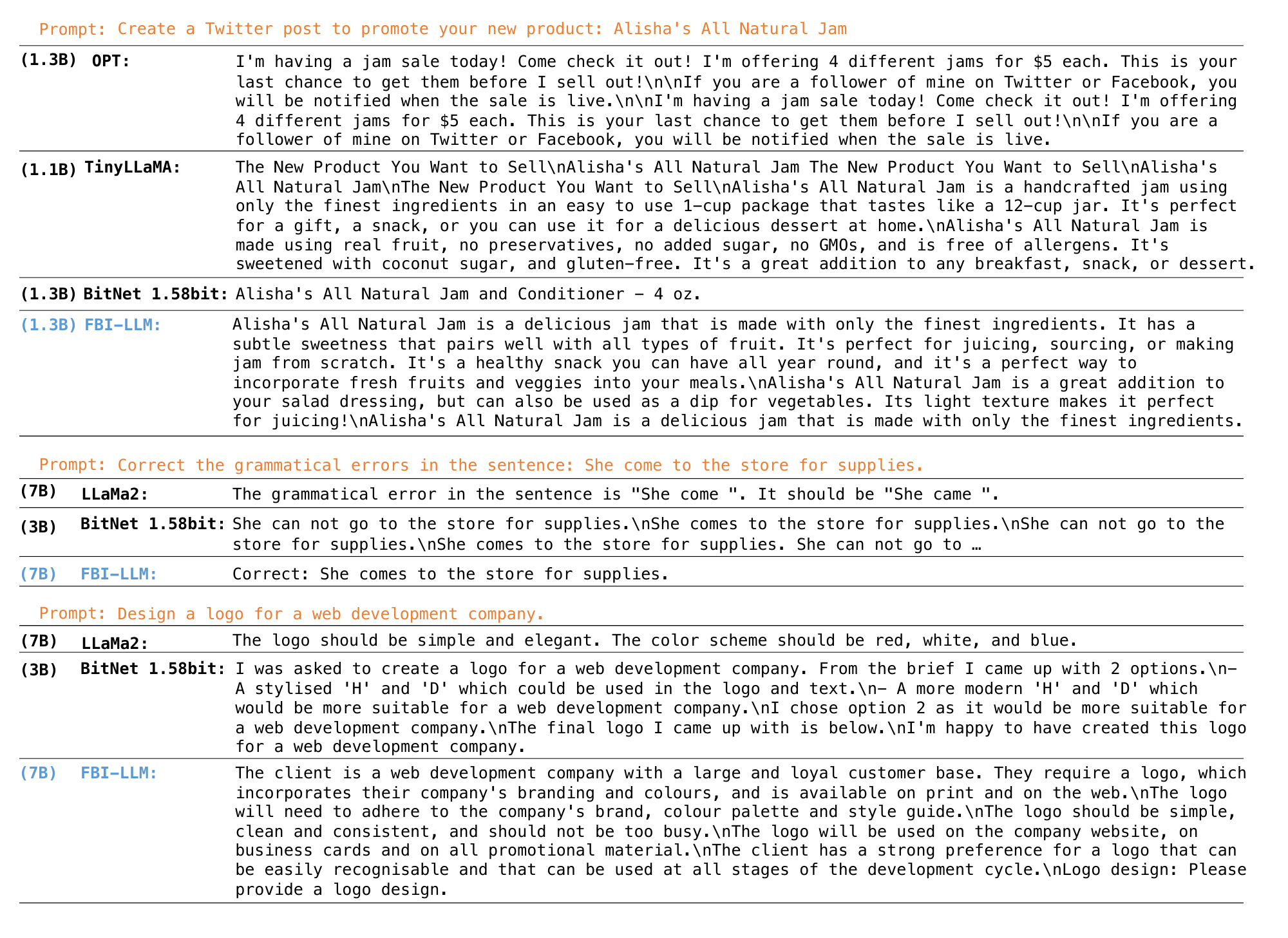}
    \caption{Generation cases. We compare the outputs of the full-precision model, BitNet b1.58, and our FBI-LLM when given the same prompts.}
    \label{fig:fbi-llm-response}
\end{figure}

As illustrated in Fig.~\ref{fig:fbi-llm-response}, although the generation quality of FBI-LLMs does not fully match that of full-precision LLMs, FBI-LLMs can still generate fluent and meaningful content. Compared to BitNet b1.58, which has a higher parameter bit-width, FBI-LLMs demonstrate a better understanding of prompts and include more knowledge in some generated examples. This indicates that FBI-LLMs possess strong generative capabilities and contain sufficient knowledge. Furthermore, FBI-LLMs demonstrate the potential to scale up further, reaching new levels of intelligence while being more hardware-friendly for deployment.

\section{Conclusion} \label{conclusion}

We have proposed a learning framework using autoregressive distillation for 1-bit weight binarization of LLMs from scratch. Extensive experiments on models of various sizes of 130M, 1.3B, and 7B demonstrate that FBI-LLM outperforms strong baselines and strikes a good balance between model size and performance. We also analyze the capabilities, properties, and potential of these extremely low-bit models, providing guidance for future research.

\noindent{\bf Limitations.} Our proposed binarization framework significantly reduces the memory and computation consumptions of LLMs, providing potential for their efficient deployment. However, there are several limitations of our models. Firstly, our 1-bit binarization inevitably incurs a performance loss compared to the original full-precision model. Additionally, the training process, which includes knowledge distillation, brings additional computational costs. Moreover, due to the unique nature of binarization, current hardware makes it difficult to directly support binarized LLMs to achieve real speedup. We also have not yet considered intermediate activation binarization which is the same as previous studies. Finally, potential ethical issues of pretrained LLMs, such as harmful biases, privacy concerns, and the spread of disinformation, are likely to persist after binarization in our LLMs.

\bibliographystyle{unsrt}
\bibliography{my.bib}


\clearpage
\appendix

\section*{Appendix}

\section{Broader Impacts}
\label{sec:broader}

Our proposed fully binarized large language models (FBI-LLM) require less computational power and memory in training and inference, making advanced AI technology accessible to organizations and individuals with limited resources. With reduced hardware requirements, smaller businesses, educational institutions, and non-profit organizations can implement LLMs, democratizing access to cutting-edge AI. Binarized models are more energy-efficient, which can significantly lower the carbon footprint associated with running large-scale AI applications. However, even binarized LLMs can still inherit and exist biases present in training data, leading to unfair outcomes in applications like hiring, law enforcement, and lending.

\section{Average Bit-width of Binarized LLM}
\label{sec:bw}
This section explains how to calculate the average bit-width of a binarized LLM. 
Since the embedding layer and head have a large number of parameters and are not binarized, we do not consider them when calculating the average bit-width. 
Consider a module containing an RMSNorm and a linear layer with a parameter matrix $A \in \mathbb{R}^{n\times n}$, which in total has $n+n^2$ parameters. 
Using the FBI-LLM binarization process as an example, after binarization, this module gains additional learnable scale parameters $\boldsymbol{\alpha}$ and $\boldsymbol{\beta}$, totaling $2n$ parameters. 
During inference, the parameters of RMSNorm, $\boldsymbol{\alpha}$, and $\boldsymbol{\beta}$ remain at 16 bits, while $A$ is quantized to 1 bit. 
Therefore, the average bit-width of this module can calculated as follows:
\begin{equation}
    \label{abw}
    \text{Average Bit-width} = \frac{1 \times n^2 + 16 \times 3n}{3n+n^2}
\end{equation}

\section{Model Configuration and Training Details}
\label{sec:confanddetail}
In this section, we list the model configurations and training details for three scales of FBI-LLM models we trained in Table~\ref{tab:config}.

\begin{table}[h]
  \caption{The configuration and training details for FBI-LLM.}
  \label{tab:config}
  \centering
  \scalebox{0.9}{
  \begin{tabular}{lccc}
    \toprule
                            &FBI-LLM 130M   &FBI-LLM 1.3B  &FBI-LLM 7B\\
    \midrule
    hidden size             & 768           & 2,048         & 4,096   \\
    intermediate size       & 2,048          & 5,632         & 11,008   \\    
    max sequence length     & 2,048          & 2,048         & 2,048  \\
    \# attention heads      & 12            & 32           & 32   \\
    \# hidden layers        & 12            & 24           & 32   \\
    \# key value heads      & 12            & 32           & 32   \\
    initializer range       & 0.02          & 0.02         & 0.02 \\
    vocabulary size              & 32,000         & 32,000        & 32,000   \\
    \hline
    learning rate           &3e-4           &3e-4          &3e-4 \\
    batch size (token)      &2M             &2.4M          &3.9M   \\
    teacher model           &LLaMA2-7B      &LLaMA2-7B     &LLaMA2-7B  \\
    \# GPUs (A100 80G)      &16             &16            &32  \\
    GPU hours for each data chunk  &130h    &189h          &729h \\
    training speed (tokens/s/GPU)  &7,800    &5,300          &1,200 \\

    \bottomrule
  \end{tabular}
  }
\end{table}

\section{Detail Experiment Results}
\label{sec:detail_exp}

We list the detailed experiment results about the effectiveness of autoregressive distillation in Section \ref{sec:ead} in Table~\ref{tab:detal}.

\begin{table}[h]
  \caption{Performance on down stream tasks and perplexity for different training objectives. Here, NA means normal autoregressive training objective and AD means autoregressive distillation training objective. Based on the number of training chunk, the table is divided into several blocks. In each block, the values with an underline represent the best value.}
  \label{tab:detal}
  \centering
  \scalebox{0.76}{
  \begin{tabular}{c|c|cccccccc|ccc}
\toprule
\multirow{2}{*}{Training Chunk}  &\multirow{2}{*}{Loss Type}  &\multicolumn{8}{c}{Zero-shot Accuracy  $\uparrow$}  &\multicolumn{3}{|c}{Perplexity $\downarrow$} \\ 
&                      &BoolQ  &PIQA  &HS    &WG    &ARC-e  &ARC-c  &OBQA  &Ave.  &Wiki2  &PTB     &C4   \\
\midrule
\multirow{2}{*}{1} &NA &42.2 &53.9 &26.2 &\underline{52.3} &\underline{29.3} &21.3 &\underline{25.4} &35.8 &85.9 &204.4 &63.9 \\ 
&AD &\underline{50.1} &\underline{54.1} &\underline{26.6} &51.0 &29.2 &\underline{21.9} &24.4 &\underline{36.8} &\underline{81.2} &\underline{193.4} &\underline{61.9} \\ 
\hline
\multirow{2}{*}{2} &NA &60.8 &\underline{56.7} &27.0 &\underline{52.2} &\underline{31.9} &20.2 &\underline{23.2} &\underline{38.9} &\underline{42.9} &\underline{128.8} &37.5 \\ 
&AD &\underline{61.5} &56.1 &\underline{27.3} &50.8 &31.1 &\underline{20.6} &22.8 &38.6 &43.4 &137.9 &\underline{37.3} \\ 
\hline
\multirow{2}{*}{3} &NA &62.0 &57.1 &27.5 &\underline{51.3} &31.5 &21.2 &24.0 &39.2 &36.7 &170.2 &\underline{32.8} \\ 
&AD &\underline{62.2} &\underline{58.1} &\underline{27.7} &51.1 &\underline{33.0} &\underline{21.3} &\underline{25.4} &\underline{39.8} &\underline{34.9} &\underline{145.2} &32.9 \\ 
\hline
\multirow{2}{*}{4} &NA &61.0 &\underline{57.9} &27.6 &\underline{52.7} &32.6 &21.0 &23.4 &39.5 &34.3 &159.6 &31.3 \\ 
&AD &\underline{62.0} &57.7 &\underline{27.7} &50.1 &\underline{33.0} &\underline{21.2} &\underline{26.8} &\underline{39.8} &\underline{32.9} &\underline{142.1} &\underline{31.0} \\ 
\hline
\multirow{2}{*}{5} &NA &61.9 &57.6 &27.7 &49.6 &32.7 &21.2 &23.6 &39.2 &32.3 &157.9 &29.7 \\ 
&AD &\underline{62.2} &\underline{58.4} &\underline{28.0} &\underline{50.8} &\underline{32.9} &\underline{21.3} &\underline{25.8} &\underline{39.9} &\underline{31.7} &\underline{137.0} &\underline{29.4} \\ 
\hline
\multirow{2}{*}{6} &NA &62.1 &\underline{59.5} &\underline{27.9} &\underline{53.1} &32.9 &21.8 &24.0 &40.2 &32.4 &147.4 &29.6 \\ 
&AD &\underline{62.2} &59.2 &27.7 &49.9 &\underline{33.9} &\underline{22.9} &\underline{26.0} &\underline{40.2} &\underline{31.3} &\underline{129.0} &\underline{29.4} \\ 
\hline
\multirow{2}{*}{7} &NA &62.0 &59.5 &27.8 &\underline{52.4} &33.8 &20.9 &\underline{25.6} &40.3 &\underline{30.0} &155.1 &\underline{28.5} \\ 
&AD &\underline{62.1} &\underline{59.5} &\underline{28.3} &51.2 &\underline{33.8} &\underline{21.8} &25.4 &\underline{40.3} &31.0 &\underline{119.3} &28.8 \\ 
\hline
\multirow{2}{*}{8} &NA &61.6 &58.8 &\underline{28.3} &\underline{51.0} &\underline{33.6} &\underline{21.3} &23.4 &\underline{39.7} &\underline{29.5} &140.0 &\underline{28.0} \\ 
&AD &\underline{62.2} &\underline{59.0} &28.1 &50.8 &32.3 &20.6 &\underline{24.0} &39.6 &30.1 &\underline{113.8} &28.9 \\ 
\hline
\multirow{2}{*}{9} &NA &60.9 &58.9 &28.2 &51.7 &\underline{34.6} &20.0 &23.4 &39.7 &\underline{29.7} &\underline{129.3} &28.4 \\ 
&AD &\underline{62.2} &\underline{60.1} &\underline{28.2} &\underline{51.8} &33.5 &\underline{21.1} &\underline{26.6} &\underline{40.5} &29.9 &129.8 &\underline{27.9} \\ 
\hline
\multirow{2}{*}{10} &NA &61.8 &\underline{59.5} &27.9 &50.3 &34.0 &20.8 &25.8 &40.0 &\underline{29.5} &144.7 &28.1 \\ 
&AD &\underline{62.2} &58.5 &\underline{28.0} &\underline{51.4} &\underline{34.2} &\underline{21.6} &\underline{26.2} &\underline{40.3} &30.1 &\underline{122.8} &\underline{28.1} \\ 
\hline
\multirow{2}{*}{11} &NA &62.1 &\underline{59.9} &\underline{28.4} &\underline{52.7} &\underline{34.1} &\underline{21.6} &\underline{25.4} &\underline{40.6} &\underline{28.8} &138.2 &\underline{27.5} \\ 
&AD &\underline{62.2} &59.0 &28.0 &49.9 &34.0 &20.7 &25.2 &39.9 &29.0 &\underline{119.9} &27.6 \\ 
\hline
\multirow{2}{*}{12} &NA &\underline{62.2} &59.1 &27.9 &51.9 &\underline{34.6} &21.2 &\underline{24.8} &40.2 &29.1 &150.3 &\underline{27.2} \\ 
&AD &62.1 &\underline{59.2} &\underline{28.1} &\underline{52.3} &34.0 &\underline{21.2} &24.2 &\underline{40.2} &\underline{28.6} &\underline{113.0} &27.3 \\ 
\hline
\multirow{2}{*}{13} &NA &\underline{62.2} &58.6 &27.9 &\underline{49.6} &\underline{34.6} &\underline{22.6} &\underline{25.4} &\underline{40.1} &28.7 &135.0 &\underline{27.0} \\ 
&AD &62.1 &\underline{58.8} &\underline{28.2} &49.4 &33.5 &21.6 &25.0 &39.8 &\underline{28.4} &\underline{134.0} &27.3 \\ 
\hline
\multirow{2}{*}{14} &NA &\underline{62.2} &58.3 &\underline{28.4} &49.4 &\underline{34.3} &20.7 &23.8 &39.6 &28.6 &138.5 &\underline{26.9} \\ 
&AD &61.5 &\underline{58.5} &28.3 &\underline{51.0} &33.9 &\underline{21.2} &\underline{26.0} &\underline{40.1} &\underline{28.3} &\underline{123.6} &27.3 \\ 
\hline
\multirow{2}{*}{15} &NA &62.2 &58.8 &\underline{28.1} &51.5 &34.1 &\underline{21.3} &\underline{26.4} &40.4 &\underline{28.0} &138.9 &\underline{26.9} \\ 
&AD &\underline{62.2} &\underline{58.8} &28.0 &\underline{51.5} &\underline{34.8} &20.9 &26.2 &\underline{40.4} &28.5 &\underline{138.3} &27.4 \\ 
\hline
\multirow{2}{*}{16} &NA &61.3 &58.4 &\underline{28.3} &51.7 &\underline{34.6} &21.7 &\underline{25.8} &40.3 &28.6 &151.9 &\underline{27.0} \\ 
&AD &\underline{62.1} &\underline{60.2} &28.1 &\underline{51.9} &33.2 &\underline{22.3} &25.2 &\underline{40.4} &\underline{28.2} &\underline{124.6} &27.2 \\ 
    \bottomrule
  \end{tabular}
}
\end{table}

\end{document}